\documentclass[letterpaper, 10 pt, conference]{ieeeconf}  

\IEEEoverridecommandlockouts                              

\usepackage{graphicx} 
\usepackage{times} 
\usepackage{amsmath} 
\usepackage{amssymb}  
\usepackage{xspace}
\usepackage{xcolor}
\usepackage{multirow}
\usepackage{booktabs}
\usepackage{caption}
\usepackage{subcaption}
\usepackage{booktabs}
\usepackage{hyperref}
\newcommand{\method}{CLAD\xspace}

\title{\LARGE \bf
\method {}: Constrained Latent Action Diffusion\\for Vision-Language Procedure Planning
}

\author{Lei Shi$^{1}$ and Andreas Bulling$^{2}$
\thanks{*Lei Shi was supported by the Wallenberg AI, Autonomous
Systems and Software Program (WASP) funded by the Knut and Alice
Wallenberg Foundation.}
\thanks{$^{1}$Lei Shi, Machine Perception and Interaction group, Örebro University, Sweden,
        {\tt\small lei.shi@oru.se}}%
\thanks{$^{2}$Andreas Bulling, Collaborative Artificial Intelligence group, University of Stuttgart,
        Germany,
        {\tt\small andreas.bulling@vis.uni-stuttgart.de}}%
}

\begin{document}

\maketitle
\thispagestyle{empty}
\pagestyle{empty}

\begin{abstract}
We propose \method{}, a \underline{C}onstrained \underline{L}atent \underline{A}ction \underline{D}iffusion model for vision-language procedure planning, the challenging task of predicting a sequence of steps that lead from a start state towards an intended goal state.
Procedure planning, while critical in robot skill learning and for assistive robots, has been largely neglected so far, and existing methods have not leveraged semantic information for step generation.
In contrast, CLAD exploits the fact that the latent space of diffusion models trained for procedure planning contains rich semantic information.
Our method uses a Variational Autoencoder (VAE) to learn the latent representation of steps and observations as constraints and integrate them into a diffusion process.
As such, our method uses these latent constraints to steer the diffusion model to generate better steps in the procedural plan.
We report extensive experiments on four datasets: three covering human procedure planning and one robot learning, and show that our method outperforms state-of-the-art methods by a large margin.
We demonstrate that the proposed integration of the step and observation representations learnt in the VAE latent space is key to these performance improvements. Code is available at \hyperlink{https://github.com/leishi07/clad}{https://github.com/leishi07/clad}.
\end{abstract}

\section{Introduction}
\label{sec:intro}

It can be expected that future human-robot interaction (HRI) scenarios will require robots that can assist humans in completing their tasks \cite{shi2021gazeemd}, whether assembly \cite{sener2022assembly101}, household \cite{padmakumar2022teach}, or daily life tasks \cite{grauman2022ego4d}.
These robots also must behave and interact with humans in an intelligent and natural manner \cite{shi2019application,sermanet2024robovqa}. 
A typical future scenario involves a human user who intends to perform a task but does not know the individual steps required to successfully complete it. 
The user could ask the robot assistant, which, in turn, would explain the intermediate steps necessary to complete the task.
Beyond providing the intermediate steps, future HRI demands that robots can also execute (some of) the planned steps to assist human users. 
This requires robots to learn the skills to perform the steps. However, it is challenging to learn skills directly from tasks, especially tasks with a long planning horizon \cite{wang2025explicit}.

Procedure planning involves generating intermediate steps given only the start and goal state of a task. 
It has significant potential to address both challenges simultaneously \cite{sun2022plate}: 
For interacting with humans, it provides the backbone for interpretable instruction and guidance. For robotic skill learning, a procedure planning method could not only decompose a complex and long-horizon task into a series of steps/subgoals that can be learned efficiently but also allows robots to select one step/subgoal to collaborate with humans. This dual utility makes procedure planning a cornerstone for advancing both interactive and autonomous robotic capabilities.
One crucial aspect to ensure natural and intelligent interaction with humans in procedure planning is that the robot needs to understand the language in interaction. This requires robots to be able to process not only visual observation but also language descriptions.
However, previous works on procedure planning use only visual observation, which makes robots lack the ability of processing language and hence limiting natural interaction with humans. 

\begin{figure}[t]
    \centering
    \includegraphics[width=0.9\linewidth]{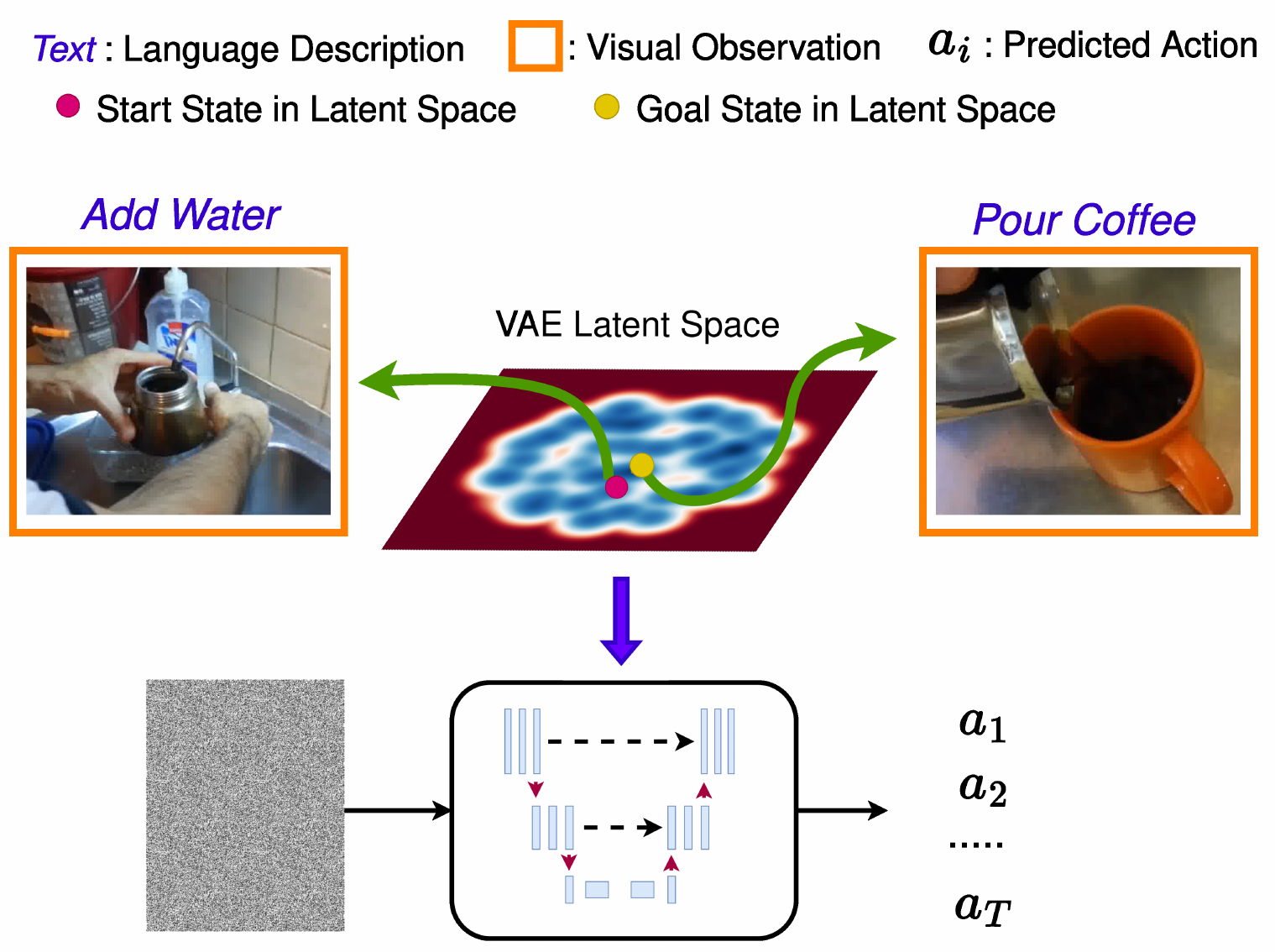}
    \caption{\method is a novel method for predicting intermediate steps in procedure planning tasks using vision-language input. 
    The start and goal steps are first encoded into a VAE latent space as constraints and then integrated into the latent space of a diffusion model to steer the action generation.
    }
    \label{fig:teasor}
\end{figure} 

To overcome this limitation, we propose vision-language procedure planning, a new procedure planning task that allows models to process both visual observation and language description of the start and goal states.
Furthermore, we propose \method{}: a \underline{C}onstrained \underline{L}atent \underline{A}ction \underline{D}iffusion model for vision-language procedure planning.
Diffusion models have shown remarkable success in various domains,
including procedure planning \cite{wang2023pdpp, shi2024actiondiffusion, nagasinghe2024not}.
However, no previous work explored the idea of \textit{using the start and goal states to constrain the predicted action sequence in the latent space}.
The latent space of diffusion models correlates primarily with the noisy input and its denoising process. 
It does not contain information on how to denoise with constraints. 
To address this limitation, we use a variational autoencoder (VAE) to learn the start and goal as constraints in its latent space:
\method first uses a VAE to learn the latent embeddings of paired visual observations and language descriptions that implicitly constrains the start and goal states. 
These learnt constraints are then injected into the deepest layer of the denoising neural network to guide the model in generating plausible action sequences.

We report experiments to evaluate \method{} on three instructional video datasets (CrossTask \cite{zhukov2019cross}, Coin \cite{tang2019coin}, NIV \cite{alayrac2016unsupervised}) and one robot learning dataset (FMB \cite{luo2025fmb}).
We opted for these datasets because they are commonly used in procedure planning and because they cover both robot assistance for HRI and robot skill learning scenarios.
For the evaluation on the instructional video datasets, we compare our proposed method with different baselines originally developed for the single-modal procedure planning task.
To make the comparison with baselines as fair as possible, we provide the ground truth actions to these baselines at inference to compensate for their inability to handle start and goal language descriptions.
Our experimental results show that our method achieves new state-of-the-art performances on all three datasets and outperforms all baselines by a large margin.
We also show experimentally that using the latent constraints learnt by the VAE allows the diffusion model to generate better action sequences. 
On the robot learning dataset, our method outperforms all baselines in all metrics. The proposed \method has perfect or near-perfect performance for shorter horizons, underlining the effectiveness of \method in breaking down long-horizon tasks in robot learning applications.  

Our specific contributions are the following:
First, we introduce the vision-language procedure planning task as a challenging and more practically useful task for real-world interactions. In contrast to previous works, this new task requires models to plan action sequences by combining visual observations with language descriptions, allowing more natural interactions with humans.
Second, we propose the first method specifically geared to vision-language procedure planning.
Our method incorporates the start and goal state representations in the latent space of the diffusion model to constrain the learning process.
Third, we evaluate our method on both instructional video datasets and robot learning datasets and show significant performance improvements over SOTA baselines, even if these baselines are given access to ground truth actions.

\section{Related Work}
\label{sec:related_work}

\subsection{Instructional Video Tasks}

Instructional videos contain multiple steps to show how to complete different tasks. A variety of research lines focused on instructional video-related work. 
Action segmentation predicts the segments of actions of given videos \cite{xu2021videoclip, zala2023hierarchical, shen2024progress}. 
A similar line of work is action step localization \cite{xu2021videoclip, dvornik2022flow, mavroudi2023learning, dvornik2023stepformer}, where the task is to align the frames in untrimmed videos with steps. 
Action recognition, unlike action segmentation and step localization, predicts the action class label of a video clip correlated to an action \cite{wang2018temporal, yang2020temporal, siddiqui2024dvanet, peng2024referring}.
Keystep recognition also predicts the action label but it uses unannotated and narrated videos and keystep vocabulary for prediction  \cite{elhamifar2020self, lin2022learning, shah2023steps, ashutosh2024video}.
Anticipation and forecasting tasks predict future actions based on observed actions in videos \cite{furnari2020rolling, gong2022future, manousaki2023vlmah, nawhal2022rethinking}.
Different from these works, the task of our work is procedure planning, which generates a sequence of actions given the start observation and the goal observation. 

\subsection{Procedure Planning}
Chang \textit{et.al} \cite{chang2020procedure} defined the procedure planning task and used MLPs and RNNs to model the dynamics between actions and observations. A search algorithm was then used to plan the intermediate actions. Later works followed the same direction by incorporating transformers \cite{sun2022plate} and Reinforcement Learning \cite{bi2021procedure}. Instead of the searching algorithms, a GAN-based method \cite{zhao2022p3iv} was used together with transformers to generate action plans. 
With the success of diffusion models, a line of works focused on using diffusion models for procedure planning. In \cite{wang2023pdpp}, the authors used the denoising diffusion probabilistic model to plan the actions by combining the start and goal observation and the task class. \cite{shi2024actiondiffusion} further used action embeddings to enrich the information of temporal dependencies between actions in the noise-adding stage.
\cite{nagasinghe2024not} constructed a knowledge graph between actions to generate action plans in combination with a diffusion model. 

\subsection{Robot Policy Learning}
Learning policies for long-horizon tasks is a key challenge in robotics \cite{mishra2023generative}, as directly training end-to-end controllers on extended sequences often leads to poor sample efficiency and suboptimal performance.
Previous works have adopted hierarchical framework to predict subgoals, however, they either only predict immediate next subgoal \cite{hatch2024ghil, shentu2024llms} or limits the possible subgoals and needs access to training examples to retrieve list of subgoals \cite{song2023llm}.
Our method \method, on the other hand, predicts the subgoals given the start and goal. Moreover, \method does not need lists of subgoals to retrieve from and does not need to limit the possible actions, showing more robustness and generalization ability. 

\section{Method}
\label{sec:method}

\subsection{Problem Definition}
\label{subsec:problem_form}
We define the vision-language procedure planning task as follows.
Given the start visual observation $o_s$, the start natural language description (short phrase) $N_s$, the goal visual observation $o_g$, and the goal natural language description (short phrase) $N_g$, the model is tasked to predict the intermediate actions that bring the start state to the goal state, i.e., the step plan $\pi=a_{1:T}$ for the time horizon $T$ and $a_{1:T}$ are categorical labels.
Here we follow the definition of action in procedure planning, i.e. an action is in the form of categorical label.
More formally, the task can be written as $p(\pi \mid o_s, o_g, N_s, N_g)$.
Our method further uses the task class $\hat{c}$ as auxiliary, hence
\begin{equation}\label{eq:problem_formation}
    p(\pi \mid o_s, o_g, N_s, N_g) = p(\pi \mid o_s, o_g, N_s, N_g, \hat{c}) \; p(\hat{c} \mid o_s, o_g),
\end{equation}
The planning first predicts the task class $\hat{c}$ using the start observation $o_s$ and the goal observation $o_g$ and then generates the action plan $\pi$ by $o_s$, $o_g$,  $N_S$, $N_g$, and $\hat{c}$.

\subsection{LatentActDiffusion}
\begin{figure*}[t]
    \centering
    \includegraphics[width=0.8\linewidth]{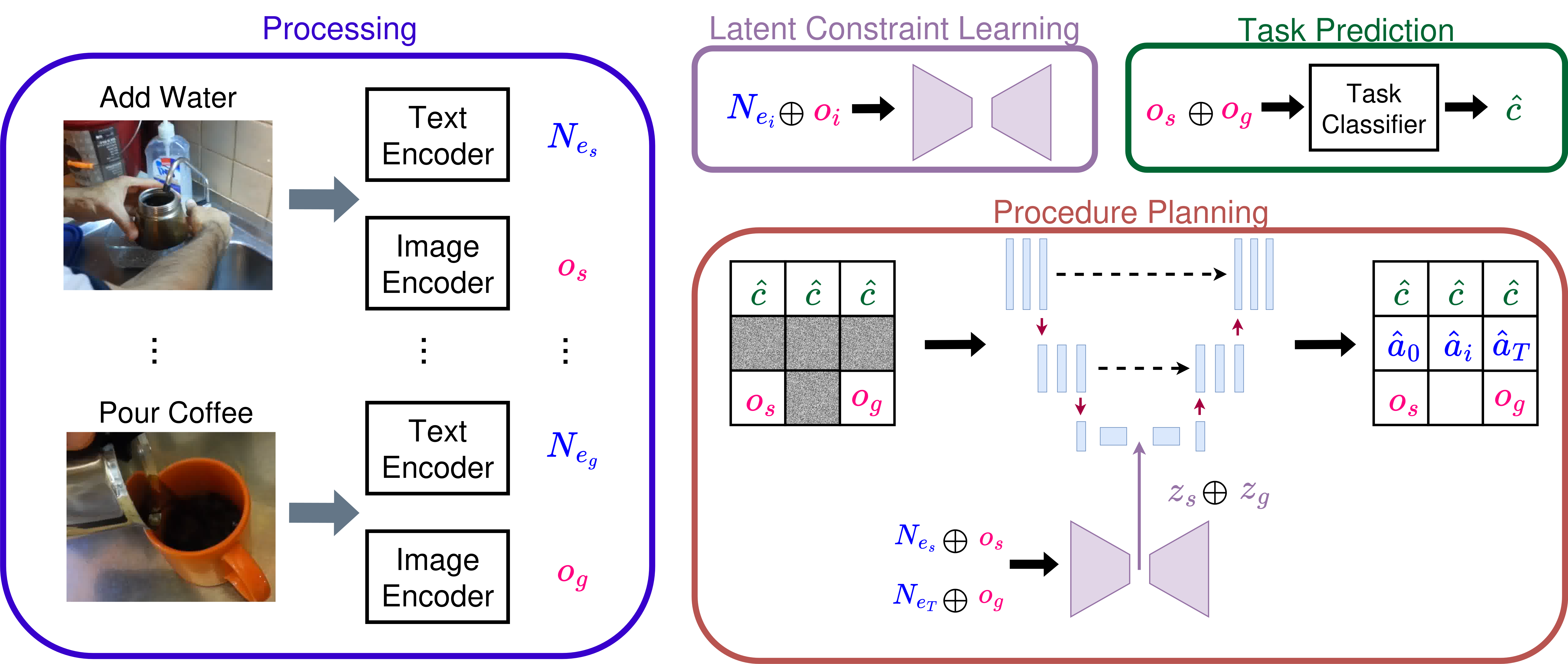}
    \caption{Overview of \method. \textbf{Processing:} We extract features from the natural language input and visual input using pretrained text encoder and image encoder. \textbf{Latent Constraint Learning:} The text feature and image feature are concatenated as the input for the VAE. \textbf{Task Prediction:} The visual features of the start and goal are used to predict the task class. \textbf{Procedure Planning:} The diffusion model task start and goal visual features and predicted task class as input to generate action sequence. The learnt latent constraints from VAE are integrated into the deepest layer of U-Net. 
    } 
    \label{fig:method}
\end{figure*}
Fig. \ref{fig:method} shows an overview of \method.
The method first encodes the input using a text encoder to extract text features from the language descriptions and an image encoder to extract features from the video frames.
We use the text and image encoder from S3D \cite{miech2020end}, following previous works \cite{wang2023pdpp, shi2024actiondiffusion, zhao2022p3iv}.
Next, we train a VAE to learn the constraints for the action sequences in latent space and train a task classifier following \cite{wang2023pdpp}. 
Lastly, we train the procedure planning model with the learnt constraints.

\subsection{Diffusion Model}
As pointed out in \cite{chang2020procedure}, performing an action will cause a change in the environment.
Essentially, the sequence of actions in an instructional video brings the start observation to the goal observation. 
The procedure planning task involves predicting the intermediate action sequence between the start and goal states.
If we interpret this action sequence as a ``path'', the start and goal states constrain this ``path'', i.e., limit what could be considered a valid path from start to goal state. 
We propose to exploit this constraint directly in the model design to improve procedure planning performance.
In \method, we use a VAE to learn the constraints in the latent space and later integrate it into the diffusion process.
The input to the VAE is the concatenation of $o_i$ and $N_{e_i}$.

\subsubsection{Procedure Planning}
Our method builds on the DDPM  \cite{ho2020denoising}.
A diffusion model incrementally adds Gaussian noise to its input $x_0$ and uses a denoising neural network to reconstruct the input from the noise.
The noise-adding process $q(x_n \mid x_{n-1})$ for $n = N, \ldots, 1$ can be described by,
\begin{equation}
    q(x_n \mid x_{n-1})=\mathcal{N}(x_n;\sqrt{1-\beta_n}x_{n-1}, \beta_n\mathbf{I}), \label{eq:q_sample}
\end{equation} 
where $x_n$ is the noised input $x_0$ after $n$ actions.
$\beta_n \in (0,1)$ is pre-defined and decides how much noise is added to $x_n$. 
After $N$ actions, $x_N$ approaches a Gaussian distribution.
In the denoising process, the model samples $x_N$ from Gaussian noise $\mathcal{N}(0, \mathbf{I})$ and denoises $x_N$ to obtain $x_0$:
\begin{equation}\label{eq:unet}
    p_\theta(x_{n-1} \mid x_n) = \mathcal{N}(x_{n-1};\mu_\theta(x_n,n),\Sigma_\theta(x_n,n)),     
\end{equation}
where $\mu_\theta(x_n,n)$ is parametrised by a neural network $\epsilon_\theta(x_n,n)$, and $\Sigma_\theta(x_n,n)$ is calculated by using $\beta_n\mathbf{I}$.
The neural network used here is a U-Net. 
Following \cite{wang2023pdpp,shi2024actiondiffusion}, we construct the input $x_0$ as follows,
\begin{equation}
x_0 = 
    \begin{bmatrix}
        c & c & ... & c & c \\
        a_0 & a_1 & ... & a_{T-1} & a_T \\
        o_s & 0 &... &0 & o_g
    \end{bmatrix},
\end{equation}
where $a_i$ is the action and $c$ is the task class.
$c$ is replaced by $\hat{c}$, the predicted task class during inference. 

\subsection{Incorporating Constraints}
A key idea of our method is to integrate the constraints learnt in the VAE space into the latent space of the diffusion model to constrain the action sequence prediction.
We opt VAE as it can learn low dimensional representations.
\cite{kwon2022diffusion} showed that the latent space of DDPM, i.e. the deepest layer in the U-Net, already contains semantic meaning. 
Inspired by this, we inject the constraints into the deepest layer in the U-Net so that the learning of denoising is implicitly constrained by the latent encodes from the VAE. Formally, we use the trained VAE to obtain the latent codes of the start $x_s$ and goal states $x_g$.
First, we form the input to the VAE,
\begin{equation}
\begin{split}
    x_{s} = o_s \oplus N_{e_s}
    \\
    x_{g} = o_g \oplus N_{e_g},
\end{split}
\end{equation}
where $N_{e_s}$ and $N_{e_g}$ are the text embeddings of start and goal language description, and $\oplus$ is concatenation. Then, we obtain the parametrised latent codes of the start and goal states,
\begin{equation}
\begin{split}
    \mathcal{N}(\mathbf{z_s}; \mathbf{\mu_s}, \mathbf{\sigma_s^2}\mathbf{I}) = q_{\phi}(\mathbf{z_s} \mid x_{s}) 
    \\
    \mathcal{N}(\mathbf{z_g}; \mathbf{\mu_g}, \mathbf{\sigma_g^2}\mathbf{I}) = q_{\phi}(\mathbf{z_g} \mid x_{g}) 
    \\
    \mathbf{z_s} = \mathbf{\mu_s}+\mathbf{\sigma_s}\odot \mathbf{\epsilon}
    \\
    \mathbf{z_g} = \mathbf{\mu_g}+\mathbf{\sigma_g}\odot \mathbf{\epsilon},
\end{split}
\end{equation}
where $q_{\phi}$ is the encoder of the VAE, $\mathbf{z_s}$ and $\mathbf{z_g}$ are the codes for the start and goal states, and $\mathbf{\epsilon} \sim \mathcal{N}(\mathbf{0}, \mathbf{I})$.
We then integrate $\mathbf{z_s}$ and $\mathbf{z_g}$ into the U-net,
\begin{equation}
    \mathbf{z_c} = f(\mathbf{z_s} \oplus \mathbf{z_g}),
\end{equation}
where $f(\cdot)$ is a neural network.
The output of $f(\cdot)$, $\mathbf{z_c}$ is added to the feature of the deepest layer of U-Net.

\subsection{Training}
Training of our method consists of two phases.
We first train the VAE, then freeze the VAE weights and train the diffusion model in the second phase. 
The loss function for training VAE is,
\begin{equation}
     \mathbf{L}_v = -\mathbf{E}_{z\sim q_\phi(z\mid x)}[\log p_{\phi(x\mid z)}]+\mathbf{KL}(q_\phi(z\mid x) \| p(z)),
 \end{equation}
where the reconstruction term (first term) is a Binary Cross Entropy loss.
For training the diffusion model, we use the ground truth task class $c$ for $x_0$.
As we need the predicted task class $\hat{c}$ during inference, we use the same network in  \cite{wang2023pdpp} to obtain $\hat{c}$.
The loss function for the diffusion model is
\begin{equation}
    \mathbf{L}_d = \sum_{n=1}^N(\mu_\theta(x_n, n) - x_0)^2,
\end{equation}
where $\mu_\theta$ is the U-Net, $x_n$ and $x_0$ are the input after $N$ actions of noise-adding and input without noise.

\section{Experiments}
\label{sec:exp}

\subsection{Datasets}\label{sec:datasets}
\subsubsection{Instructional Video Dataset}

We evaluate \method on CrossTask \cite{zhukov2019cross}, Coin \cite{tang2019coin}, and NIV \cite{alayrac2016unsupervised}.
All datasets have language descriptions of actions either in the form of a verb plus a noun or short phrases. Action labels of $a_i$ are in the form of categorical labels. 
Using a sliding window, we extract action sequences with the time horizon $T$.
For an extracted action sequence $[a_i, ... , a_{i+T-1}]$, each action has a corresponding video clip.
There are two settings for extracting the video features for $o_s$ and $o_g$.
We refer to them as the PDPP \cite{wang2023pdpp} and the KEPP setting \cite{nagasinghe2024not}.
In both cases, the features were extracted by using a model \cite{miech2020end}, pre-trained on the HowTo100M dataset \cite{miech2019howto100m}. 
In the PDPP setting, the start observation of an action sequence begins at the time of the first action and ends at three seconds after the first action.
The goal observation of an action sequence begins two seconds before the last action and ends one second after the last action. 
In the KEPP setting, the start observation of an action sequence begins one second before the first action and ends two seconds after the first action. The goal observation of an action sequence begins one second before the last action and ends two seconds after the last action. 
We use KEPP setting in our experiments.
 
\subsubsection{Robot Learning Dataset}
We also evaluate \method on a robot learning dataset: Functional Manipulation Benchmark (FMB) \cite{luo2025fmb}. It is a real-world dataset designed for learning robotic manipulation skills in long-horizon tasks. The tasks are single object manipulation and multiple object manipulation in an assembly scenario. The dataset has descriptions of actions across full episodes. The descriptions are text such as
``grasp'' and ``move up''.
Next, we curate the FMB dataset for the use of vision-language procedure planning task. We split all episodes into train set and test set with the ratio of 70\% and 30\%. In each episode, we use the descriptions of the actions as the descriptions of the actions. We slice the whole sequence of actions into sequences with time horizon T, with padding if necessary. For each sliced sequence, we extract the first frame of the first action and the last frame of the last action as the start and goal observation.

\subsection{Baselines}
We compared our method to the following baselines on the instructional video datasets: random, PDPP \cite{wang2023pdpp}, ActionDiffusion \cite{shi2024actiondiffusion}, SCHEMA \cite{niu2024schema}, and KEPP \cite{nagasinghe2024not}. For random baseline, we select random actions to composite a sequence.
Since we propose a new procedure planning task and all of these methods were developed for the single-modal procedure planning task, i.e., without additional language descriptions, it was impossible to directly compare \method{} with these baselines.
To ensure a fair comparison, we therefore modified the evaluation scheme of the baseline methods:
We changed the predictions of the first action $\hat{a_0}$ and the last action $\hat{a_T}$ to the ground truth $a_0$ and $a_T$.
We trained all baseline models using the official code provided by the authors.
All baselines were trained with KEPP data curation settings. All training parameters were kept the same as in the original papers.
For the FMB dataset, we use PDPP setting and compare PDPP, ActionDiffusion, and KEPP with our proposed \method. We leave SCHEMA out since it needs detailed descriptions of actions in natural language, and FMB dataset lacks of them.

\subsection{Metrics}
As commonly done in procedure planning, we used Success Rate (SR), mean Accuracy (mAcc) and mean Single Intersection over Union (mSIoU) as the evaluation metrics.
SR evaluates if a generated action plan is successful or not.
An action plan is successful if all actions and their order are correct.
This is the strictest metric among all metrics.
mAcc is calculated based on the actions in the action plan. The order of the actions is not considered. 
mSIoU calculates the IoU of the actions and the ground truth actions in action plans. It also does not consider the order of actions.
The works in \cite{chang2020procedure, zhao2022p3iv, bi2021procedure} calculated mIoU with all action plans in a mini-batch. \cite{wang2023pdpp} showed that calculating mIoU depends on the batch size. Later works \cite{shi2023inferring} also used mSIoU.

\subsection{Computational Cost}
The number of parameters of the VAE is around 2.3M and the number of parameters of the diffusion model is around 165.6M. All experiments use one RTX 4070 Super GPU.
\begin{table*}[h]\centering
\caption{Comparison of the different methods on the CrossTask dataset with time horizons $T=3$ to $T=6$. All baselines use ground truth $a_0$ and $a_t$ for inference, ours does not use ground truth $a_0$ and $a_t$. Best results are marked in \textbf{bold}. 
}\label{tab:crosstask_sota_t34}
\resizebox{0.7\linewidth}{!}{
\begin{tabular}{lcccccc|cc}\toprule
& T=3 & & & T=4 & & & T=5 & T=6 \\\cmidrule{2-7} \cmidrule{8-9}
Method &SR↑ &mAcc↑ &mSIoU↑ &SR↑ &mAcc↑ &mSIoU↑ &SR↑ &SR↑ \\\midrule
Random &0	&1.02	&1.7	&0	&1	&1.83	&0 &0 \\
PDPP \cite{wang2023pdpp} &50.94 &83.65 &79.44 &27.86 &72.69 &74.16 &16.57 & 9.98\\
ActionDiffusion \cite{shi2024actiondiffusion} &46.35 &82.12 &77.48 &23.61 &70.64 &72.32 & 14.17 &8.13 \\
SCHEMA \cite{niu2024schema} &47.06 &82.35 &- &27.66 &71.77 &- &15.77 &10.1\\
KEPP \cite{nagasinghe2024not} &47.98 &82.66 &77.94 &26.76 &72.52 &73.22 & 16.08 & 9.59 \\
\cmidrule{1-9}
\textbf{\method{} (ours)} &\textbf{59.35} &\textbf{84.56} &\textbf{80.61} &\textbf{34.09} &\textbf{74.3} &\textbf{74.85} & \textbf{21.42} & \textbf{12.66} \\
\bottomrule
\end{tabular}
}
\end{table*}

\section{Results}
\label{sec:results}

\subsection{Performance on Instructional Video Datasets}\label{sec:sota}
\subsubsection{CrossTask}\label{sec:crosstask_sota}
Table \ref{tab:crosstask_sota_t34} shows a performance comparison of the baseline methods on the CrossTask dataset for time horizons $T=3$ to $T=6$. Following previous works \cite{wang2023pdpp, shi2024actiondiffusion}, we only report SR for $T=5$ and $T=6$.
Our method outperforms all baselines on all metrics for $T=3$ and $T=4$.
Our SRs when $T=3$ and $T=4$ are over 10\% and 7\% higher than the second-best model. mAcc and mSIoU, on the other hand, are close. The reason is that all baselines use ground truth start and goal actions during inference. It results in more correct individual actions, thus mAcc and mSIoU are higher. This also shows that our method can predict the middle actions more accurately.
For the results when $T=5$ and $T=6$,
\method has SOTA performances compared to all baselines. 

\subsubsection{Coin}
\begin{table}[!htp]\centering
\caption{Comparison with SOTA methods on Coin dataset with time horizon $T=3$ and $T=4$. 
All baselines use ground truth $a_0$ and $a_t$ for inference,  ours does not use ground truth $a_0$ and $a_t$.
Best results are marked in \textbf{bold}.
}\label{tab:coin_sota}
\resizebox{\linewidth}{!}{
\begin{tabular}{lccccccc}\toprule
& & & T=3 & & & T=4 & \\\cmidrule{2-7}
Method &SR↑ &mAcc↑ &mSIoU↑ &SR↑ &mAcc↑ &mSIoU↑ \\\midrule
Random &0	&0.11	&0.18	&0	&0.1	&0.19 \\
PDPP \cite{wang2023pdpp}  &38.73 &79.58 &75.54 &21.6 &69.46 &73.21 \\
ActionDiffusion\cite{shi2024actiondiffusion} &38.5 &79.5 &75.5 &20.63 &68.21 &72.33 \\
SCHEMA \cite{niu2024schema}  &46.82 &\textbf{82.27} &- &30.08 &70.97 &- \\
KEPP \cite{nagasinghe2024not} &36.87 &78.96 &73.91 &21.15 &67.94 &71.11 \\
\cmidrule{1-7}
\textbf{\method{} (ours)} &\textbf{60.75} &79.58 &\textbf{79.07} &\textbf{42.15} &\textbf{73.01} &\textbf{77.17} \\
\bottomrule
\end{tabular}
}
\end{table}

Table \ref{tab:coin_sota} shows the results on the Coin dataset.
\method outperforms all baselines except SCHEMA on all metrics. Although SCHEMA achieves higher mAcc, the SR of \method surpasses SCHEMA by large margins. The reason is that the start and goal actions are ground truth actions during the inference for SCHEMA, while ours do not use ground truth actions. 

\subsubsection{NIV}
\begin{table}[!htp]\centering
\caption{Comparison with SOTA methods on NIV dataset with time horizon $T=3$ and $T=4$. 
All baselines use ground truth $a_0$ and $a_t$ for inference,  ours does not use ground truth $a_0$ and $a_t$.
Best results are marked in \textbf{bold}.
}\label{tab:niv_sota}
\resizebox{\linewidth}{!}{
\begin{tabular}{lccccccc}\toprule
& & & T=3 & & & T=4 & \\\cmidrule{2-7}
Method &SR↑ &mAcc↑ &mSIoU↑ &SR↑ &mAcc↑ &mSIoU↑ \\\midrule
Random &0	&2.59	&3.44	&0	&1.86	&3.41 \\
PDPP \cite{wang2023pdpp} &40.37 &80.12 &75 &27.51 &69.65 &71.38 \\
ActionDiffusion \cite{shi2024actiondiffusion} &37.41 &79.14 &73.46 &19.65 &63.76 &65 \\
SCHEMA \cite{niu2024schema} &41.44 &80.48 &- &30.90 &68.84 &- \\
KEPP \cite{nagasinghe2024not} &39.63 &79.88 &74.01 &26.64 &67.90 &71.37 \\
\cmidrule{1-7}
\textbf{\method{} (ours)}  &\textbf{70.74} &\textbf{84.81} &\textbf{86.98} &\textbf{56.33} &\textbf{78.82} &\textbf{84.71} \\
\bottomrule
\end{tabular}
}
\end{table}

Table \ref{tab:niv_sota} shows the results on the NIV dataset. We have SOTA performance on all metrics over $T=3$ and $T=4$. We outperform other baselines on SR by a large margin. The likely reason for this is the small dataset size of NIV. The number of curated training samples is less than 700, and the transformer-based baseline (SCHEMA) and diffusion model-based baselines (PDPP, ActionDiffusion, KEPP) struggle with the small dataset. 
In contrast, while our method also uses a diffusion model, thanks to the latent constraints, our SRs are around 30\% higher than the second-best SRs.
This finding underlines the effectiveness of the proposed integration of the VAE latent constraints.

\subsection{Performance on FMB Dataset}

\begin{table*}[h]
\caption{Comparison with SOTA metods on the FMB dataset for different time horizons $T$. Higher numbers are better, best results are marked in \textbf{bold}.}
\label{tab:fmb}
\begin{tabular}{l *{4}{ccc}}
\toprule
& \multicolumn{3}{l}{T=3} & \multicolumn{3}{l}{T=4} & \multicolumn{3}{l}{T=5} & \multicolumn{3}{l}{T=6} \\
\midrule
Method & SR↑ & mAcc↑ & mSIoU↑ & SR↑ & mAcc & mSIoU↑ & SR↑ & mAcc↑ & mSIoU↑ & SR↑ & mAcc↑ & mSIoU↑ \\
\midrule
PDPP \cite{wang2023pdpp} & 92.35 & 95.87 & 95.29 & 86.56 & 92.39 & 92.43 & 85.73 & 91.82 & 95.17 & 62.67 & 79.74 & 86.44 \\
ActionDiffusion \cite{shi2024actiondiffusion} & 70.03 & 79.02 & 80.87 & 46.72 & 64.03 & 67.49 & 35.45 & 50.82 & 68.86 & 24.98 & 49.05 & 69.35 \\
KEPP \cite{nagasinghe2024not} & 92.83 & 96.35 & 95.74 & 87.09 & 92.59 & 92.61 & 85.45 & 91.55 & 94.73 & 60.57 & 79.71 & 86.91 \\
\midrule
\textbf{\method (Ours)} & \textbf{100} & \textbf{100} & \textbf{100} & \textbf{98.87} & \textbf{99.44} & \textbf{99.72} & \textbf{94.7} & \textbf{97.88} & \textbf{98.86} & \textbf{70.4} & \textbf{85.89} & \textbf{90.35} \\
\bottomrule
\end{tabular}
\end{table*}

Table \ref{tab:fmb} shows the results of \method on FMB dataset with time horizon 3 to 6. 
\method achieves SOTA in all metrics across all time horizons.
The SR, mAcc, and mSIoU all degrade as the time horizon $T$ increases. For short horizons ($T \leq 4$), \method achieves perfect or near-perfect performance (SR = 100.0\% at $T=3$, 98.9\% at $T=4$). 
When time horizon $T=6$, the SR declines sharply to 70.4\%.
PDPP and KEPP have comparable performance across all time horizons. ActionDiffusion, on the other hand, struggles the most, especially when $T$ is larger than three.

Compared to the instructional video datasets, the perfect and near-perfect performances of \method at short horizons ($T=3, 4$) are probably because of the action space of FMB is much smaller than CrossTask, Coin and NIV (7 vs 105, 778, and 48). This makes the search space for predicting action sequence much smaller. 
However, it is still a challenge for planning the actions in correct orders in longer horizons. For $T=6$, mAcc and mSIoU of \method are 85.89\% and 90.35\% while SR is 70.4\%. This indicates that the failures in SR may arise primarily from the sequential composition of actions rather than individual action accuracy.

Overall, the results strongly validate the effectiveness of \method on robotics application, i.e. planning intermediate action of a task from visual and language input. The significant performance on the FMB dataset underscores the  potential of \method as a strong method for subgoal prediction in hierarchical long-horizon planning in robotic applications. 

\subsection{Ablation Studies}
\subsubsection{Impact of Language Description}
\begin{table}[h]
    \centering
    \caption{Comparison of \method and \method using visual observation only on NIV dataset for $T=3$. $\dagger$ indicates \method using visual observation only.}
    \label{tab:visual_only}
    \begin{tabular}{l l c c c}
        \toprule
        Dataset & Method & SR$\uparrow$ & mAcc$\uparrow$ & mSIoU$\uparrow$ \\
        \midrule
        CrossTask & \method$\dagger$ & 28.34 & 56.38 & 62.12 \\
                  & \method          & \textbf{59.35} & \textbf{84.56} & \textbf{80.61} \\
        \midrule
        Coin      & \method$\dagger$ & 18.94 & 42.93 & 49.71 \\
                  & \method          & \textbf{60.75} & \textbf{79.58} & \textbf{79.07} \\
        \midrule
        NIV       & \method$\dagger$ & 24.07 & 41.23 & 53.85 \\
                  & \method          & \textbf{70.74} & \textbf{84.81} & \textbf{86.98} \\
        \bottomrule
    \end{tabular}
\end{table}

To show the impact of using language in vision-language procedure planning, we compare \method with \method only using visual observations.
Table \ref{tab:visual_only} shows the results on CrossTask, Coin and NIV dataset for $T=3$. \method outperforms \method using visual observation only on all metrics by large margin, demonstrating the importance and necessity of using language description of actions in procedure planning.

\subsubsection{VAE without Visual Observation}
\begin{table}[h]
    \centering
    \caption{Comparison of \method and \method with vae trained only with language description (\method$\dagger$).}
    \label{tab:act_only_niv}
    \begin{tabular}{lccc}
        \toprule
        & SR$\uparrow$ & mAcc$\uparrow$ & mSIoU$\uparrow$ \\
        \midrule
        \method$\dagger$ & 69.26 & 84.44 & 85.92 \\
        \method          & \textbf{70.74} & \textbf{84.81} & \textbf{86.98} \\
        \bottomrule
    \end{tabular}
\end{table}
We train the VAE with both visual observations and language descriptions for \method. We show the effect of training the VAE only with language description.
Table \ref{tab:act_only_niv} shows the results on NIV with $T=3$. Training VAE with both visual observation and language description results in better planning performance. 
It empirically shows the latent constraints containing visual and language are more compatible with the diffusion model.

\section{Conclusion}
\label{sec:conclusion}

This paper introduced vision-language procedure planning as a novel task integrating visual observations and language descriptions.
This setting is particularly important for future usage scenarios in which robots learn skills and interact with humans.
To address this novel task, we proposed \method{} which leverages a VAE to learn latent constraints and integrates them into a diffusion model for step sequence generation.
Our experiments on instructional video and robot learning datasets demonstrated significant performance improvements of \method{} over several state-of-the-art baselines, thereby underlining the effectiveness of the proposed integration of constraints into the learning problem.
One direction of future work will be focused on further improving the performances on long-horizon planning. Another direction is to incorporate re-planning mechanisms for robotics application.

\addtolength{\textheight}{-0cm}   




\bibliographystyle{IEEEtran}
\bibliography{ref.bib}

\end{document}